\documentclass[journal,10pt,twocolumn,twoside]{IEEEtran}
\bstctlcite{IEEEexample:BSTcontrol}
\usepackage{times,amsmath,amssymb}
\usepackage{graphicx}
\usepackage{setspace}
\usepackage[ruled]{algorithm2e}
\usepackage{algorithmicx}
\usepackage{algpseudocode}
\usepackage{pdfpages}

\title{Probabilistic Class-Specific Discriminant Analysis}
\author{\IEEEauthorblockN{Alexandros Iosifidis}\\
\IEEEauthorblockA{Department of Engineering, ECE, Aarhus University, Denmark\\
Email: alexandros.iosifidis@eng.au.dk}
}
\normalsize
\begin{document}
\maketitle

\newpage

\begin{abstract}
In this paper we formulate a probabilistic model for class-specific discriminant subspace learning. The proposed model can naturally incorporate the multi-modal structure of the negative class, which is neglected by existing class-specific methods. Moreover, it can be directly used to define a class-specific probabilistic classification rule in the discriminant subspace. We show that existing class-specific discriminant analysis methods are special cases of the proposed probabilistic model and, by casting them as probabilistic models, they can be extended to class-specific classifiers. We illustrate the performance of the proposed model in both verification and classification problems.
\end{abstract}

\section{Introduction}\label{S:Intro}
Class-Specific Discriminant Analysis (CSDA) \cite{arashloo2014csksr,iosifidis2015CSRDA,iosifidis2015cskernelspace,tran2017mlcsda}, determines an optimal subspace suitable for verification problems, where the objective is the discrimination of the class of interest from the rest of the world. As an example, let us consider the person identification problem, either through face verification \cite{kittler2000faceverification}, or through exploiting appearance or movement information \cite{cao2019towards,iosifidis2012TIFS}. Different from person recognition, which is a multi-class classification problem assigning a sample (facial image or movement sequence) to a class in a pre-defined set of classes (person IDs in this case), person identification discriminates the person of interest from all rest people.

While multi-class discriminant analysis models, like Linear Discriminant Analysis (LDA) and its variants \cite{duda2000pattern,iosifidis2013optimal,iosifidis2014krda,ye2017fast,souza2020enhanced,zafeiriou2012regularized} can be applied in such problems, they are inherently limited by the adopted class discrimination definition. That is, the maximal dimensionality of the resulting subspace is restricted by the number of classes, due to the rank of the between-class scatter matrix. In verification problems involving two classes LDA and its variants lead to one-dimensional subspaces. On the other hand, CSDA by expressing class discrimination using the out-of-class and intra-class scatter matrices is able to define subspaces the dimensionality of which is restricted by the the number of samples forming the smallest class (which is usually the class of interest) or the number of original space dimensions. By defining multiple discriminant dimensions, CSDA has been shown to achieve better class discrimination and better performance in verification problems compared to LDA \cite{arashloo2014csksr,iosifidis2015CSRDA}.
\begin{figure}[]
\centering \centerline{
\includegraphics[width=0.9\linewidth]{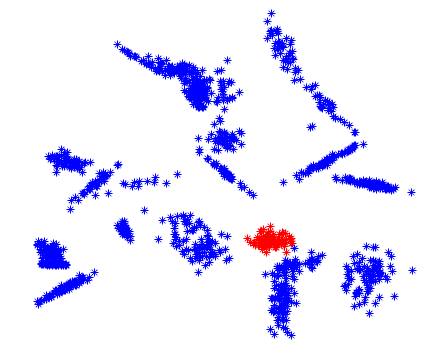} }
\caption{\it A toy example of a $2$-dimensional problem illustrating the case where data forming the class of interest (red) is surrounded by data forming other classes. In class-specific learning the labels of the the other classes are not available, leading to the assignment of the corresponding data to the negative class label (blue).}\label{fig:Example}
\end{figure}

While the definition of class discrimination in CSDA and its variants based on the intra-class and out-of-class scatter matrices overcomes the limitations of LDA related to the dimensionality of the discriminant subspace, it overlooks the structure of the negative class. Since in practice samples forming the negative class belong to many classes, different from the positive one, it is expected that they will form subclasses, as illustrated in Figure \ref{fig:Example}. Class discrimination as defined by CSDA and its variants disregards this structure. Related research in multi-class discriminant analysis indicates that exploitation of subclass information can enhance discrimination power \cite{chen2003facial,zhu2006subclass,yu2018onlinefault,iosifidis2013rcvcda}.

In this paper, we formulate the class-specific discriminant analysis optimization problem based on a probabilistic model that incorporates the above-described structure of the negative class. We show that the optimization criterion used by standard CSDA and its variants corresponds to a special case of the proposed probabilistic model, while new discriminant subspaces can be obtained by allowing samples of the negative class to form subclasses automatically determined by applying (unsupervised) clustering techniques on the negative class data. Moreover, the use of the proposed probabilistic model for class-specific discriminant learning naturally leads to a classification rule in the discriminant subspace, something that is not possible when the standard CSDA criterion is considered.

\section{Related Work}\label{S:ProblemStatement}
Let us denote by $\mathcal{S}_p = \{\mathbf{x}_1,\dots,\mathbf{x}_{N_p}\}$ a set of $N_p$ $D$-dimensional vectors representing samples of the positive class and by $\mathcal{S}_n = \{\mathbf{x}_{N_p+1},\dots,\mathbf{x}_{N}\}$, where $N = N_p + N_n$, a set of $N_n$ vectors representing samples of the negative class. In the following we consider the linear class-specific subspace learning case and we will describe how to perform nonlinear (kernel-based) class-specific subspace learning following the same processing steps in Section \ref{SS:Extensions}. We would like determine a linear projection $\mathbf{W} \in \mathbb{R}^{D \times d}$, mapping $\mathbf{x}_i$ to a $d$-dimensional subspace, i.e. $\mathbf{z}_i = \mathbf{W}^T \mathbf{x}_i$ that enhances discrimination of the two classes.

Class-specific Discriminant Analysis defines the projection matrix $\mathbf{W}$ as the one maximizing the following criterion:
\begin{equation}\label{Eq:CSKDA_J2}
\mathcal{J}(\mathbf{W}) = \frac{ Tr \left( \mathbf{W}^T \mathbf{S}_n \mathbf{W} \right) }{ Tr \left( \mathbf{W}^T \mathbf{S}_p \mathbf{W} \right) },
\end{equation}
where $Tr(\cdot)$ is the trace operator. $\mathbf{S}_n \in \mathbb{R}^{D \times D}$ and $\mathbf{S}_p \in \mathbb{R}^{D \times D}$ are the out-of-class and intra-class scatter matrices:
\begin{equation}\label{Eq:S_n_matrix}
\mathbf{S}_n = \sum_{\mathbf{x}_i \in \mathcal{S}_n} \left( \mathbf{x}_i - \mathbf{m} \right)\left( \mathbf{x}_i - \mathbf{m} \right)^T
\end{equation}
\begin{equation}\label{Eq:S_p_matrix}
\mathbf{S}_p = \sum_{\mathbf{x}_i \in \mathcal{S}_p} \left( \mathbf{x}_i - \mathbf{m} \right)\left( \mathbf{x}_i - \mathbf{m} \right)^T
\end{equation}
where $\mathbf{m}$ is the mean vector of the positive class, i.e. $\mathbf{m} = \frac{1}{N_p} \sum_{\mathbf{x}_i \in \mathcal{S}_p} \mathbf{x}_i$. $\mathbf{W}$ is obtained by solving the generalized eigen-analysis problem of $\mathbf{S}_n \mathbf{w} = \lambda \mathbf{S}_p \mathbf{w}$ and keeping the eigenvectors corresponding to the $d$ largest eigenvalues \cite{jia2009trace}. In the case where $\mathbf{S}_n$ is singular, a regularized version of the above problem is solved.

A Spectral Regression \cite{cai2007spectral} based solution of (\ref{Eq:CSKDA_J2}) has been proposed in \cite{arashloo2014csksr,iosifidis2015CSRDA}. Let us denote by $\mathbf{w}$ an eigenvector of the generalized eigen-analysis problem $\mathbf{S}_n \mathbf{w} = \lambda \mathbf{S}_p \mathbf{w}$ with eigenvalue $\lambda$. By setting $\mathbf{X}^T \mathbf{w} = \mathbf{v}$ ($\mathbf{X}$ being the data matrix), the original eigen-analysis problem can be transformed to the following eigen-analysis problem $\mathbf{P}_n \mathbf{v} = \lambda \mathbf{P}_p \mathbf{v}$, where $\mathbf{P}_n = \mathbf{e}_n \mathbf{e}_n^T - \frac{1}{N_p} \mathbf{e}_n \mathbf{e}_p^T - \frac{1}{N_p} \mathbf{e}_p \mathbf{e}_n^T + \frac{1}{N_p^2} \mathbf{e}_p \mathbf{e}_p^T$ and $\mathbf{P}_p = (1 - \frac{2}{N_p} + \frac{1}{N_p^2}) \mathbf{e}_p \mathbf{e}_p^T$. Here $\mathbf{e}_p \in \mathbb{R}^N$ and $\mathbf{e}_n \in \mathbb{R}^N$ are the positive and negative class indicator vectors. $\mathbf{W}$ is then obtained by applying a two-step process:
\begin{itemize}
    \item Solution of the eigen-analysis problem $\mathbf{P}_n \mathbf{v} = \lambda \mathbf{P}_p \mathbf{v}$ to determine the matrix $\mathbf{V} = [\mathbf{v}_1,\dots,\mathbf{v}_{d}]$, where $\mathbf{v}_i$ is the eigenvector corresponding to the $i$-th largest eigenvalue.
    \item Calculation of $\mathbf{w}_i, \:i=1,\dots,d$ such that $\mathbf{X}^T\mathbf{w}_i = \mathbf{v}_i$.
\end{itemize}
It has been shown in \cite{iosifidis2016sucskda} that the generalized eigenvectors $\mathbf{v}$ of $\mathbf{P}_p$ and $\mathbf{P}_n$ can be directly obtained using the labeling information of the training vectors. However, in that case the order of the eigenvectors is not related to their discrimination ability. It has been also shown in \cite{iosifidis2017cskdaRev} that the class-specific discriminant analysis problem in (\ref{Eq:CSKDA_J2}) is equivalent to a low-rank regression problem in which the target vectors are the same as those defined in \cite{iosifidis2016sucskda}, providing a new proof of the equivalence between class-specific discriminant analysis and class-specific spectral regression.

After determining the data projection matrix $\mathbf{W}$, the training vectors $\mathbf{x}_i, \:i=1,\dots,N$ are mapped to the discriminant subspace $\mathbf{z}_i = \mathbf{W}^T \mathbf{x}_i$. When a classification problem is considered, a classifier is trained using $\mathbf{z}_i$. For example, \cite{iosifidis2015CSRDA} trains a linear SVM on vectors $\mathbf{d}_i = |\mathbf{z}_i - \mbox{\boldmath$\mu$}|$, where $\mbox{\boldmath$\mu$} = \mathbf{W}^T \mathbf{m}$ and the absolute value operator is applied element-wise on the resulting vector. Due to the need of training an additional classifier, class-specific discriminant analysis models are usually employed in class-specific ranking settings, where test vectors $\mathbf{x}^{*}_j$ are mapped to the discriminant subspace $\mathbf{z}^{*}_j = \mathbf{W}^T \mathbf{x}^{*}_j$ and are subsequently ordered based on their distance w.r.t. the positive class mean $d_j = \|\mathbf{z}^{*}_j - \mbox{\boldmath$\mu$}\|_2$.

\section{Probabilistic Class-Specific Learning}\label{S:ProposedMethods}
In this section, we follow a probabilistic approach for defining a class-specific discrimination criterion that is able to encode subclass information of the negative class. We call the proposed method Probabilistic Class-Specific Discriminant Analysis (PCSDA). PCSDA defines a subspace $\mathbb{R}^d, \:d \le D$ of a feature space $\mathbb{R}^D$ such that the positive class is optimally discriminated from the negative class, based on the assumption that the negative class is formed by multiple subclasses having the same cardinality. We show how to relax this assumptions in the next subsection. 

Let us denote by $\mathbf{x} \in \mathbb{R}^D$ a random variable, realizations of which correspond to samples of the positive and negative classes in our problem. PCSDA assumes there exists a positive class $c_p$ following a multivariate Gaussian distribution defined by the mean vector $\mathbf{m} \in \mathbb{R}^D$ and the covariance matrix $\mathbf{\Phi}_p \in \mathbb{R}^{D \times D}$, i.e.:
\begin{equation}
P(\mathbf{x}|c_p) \sim N(\mathbf{x}|\mathbf{m},\mathbf{\Phi}_p).
\end{equation}
Since the negative class $c_n$ is formed by samples belonging to multiple classes (which are not distinguished by the labeling information available during training), PCSDA assumes that the negative class forms subclasses, each of which follows a multivariate Gaussian distribution in $\mathbb{R}^D$. Let $\mathbf{y} \in \mathbb{R}^D$ be a random variable a realization of which corresponds to the mean vector of a negative subclass. Since in class-specific learning we are interested in maximizing the scatter of the negative data from the positive class (represented by $\mathbf{m}$), we model the distribution of $\mathbf{y}$ with respect to $\mathbf{m}$ as a multivariate Gaussian distribution:
\begin{equation}
P(\mathbf{y}) \sim N(\mathbf{y} | \mathbf{m},\mathbf{\Phi_n}),
\end{equation}
where $\mathbf{\Phi_n}$ is the corresponding covariance matrix expressing the scatter of the negative subclasses with respect to the positive class mean $\mathbf{m}$. Then, the distribution of the samples from the negative subclasses with respect to the positive class is expressed by the following multi-modal distribution \cite{Mood1974introduction}:
\begin{equation}
P(\mathbf{x}|c_n) = \int N(\mathbf{y}|\mathbf{m},\mathbf{\Phi}_n) \: N(\mathbf{x}|\mathbf{y},\mathbf{\Phi}_w) \: d \mathbf{y},
\end{equation}
where $\mathbf{\Phi}_w$ is the (common) covariance matrix of the negative subclasses.

\subsection{Training phase}\label{SS:TrainingPhase}
Given a set of positive i.i.d. samples $\mathcal{S}_p = \{\mathbf{x}_i,\dots,\mathbf{x}_{N_p}\}$ the probability of correct assignment under our model is:
\begin{equation}
P(\mathcal{S}_p|c_p) = \prod_{i=1}^{N_p} P(\mathbf{x_i}|c_p). \label{Eq:P_Sp_Prod}
\end{equation}
Let us assume that negative class is formed by samples $\mathbf{x}_i, \:i=N_p+1,\dots,N$ belonging to $K$ subclasses, i.e. $\mathcal{S}_n = \{\mathcal{S}_1,\dots,\mathcal{S}_K\}$, each having a cardinality of $M = N_n / K$. The probability of assigning each of the negative samples to the corresponding negative subclass is given by:
\begin{equation}
P(\mathcal{S}_k|c_n) = \int N(\mathbf{y}|\mathbf{m},\mathbf{\Phi}_n) \: \prod_{\mathbf{x}_i \in \mathcal{S}_k} P(\mathbf{x}_i|\mathbf{y},\mathbf{\Phi}_w) \:d \mathbf{y}.  \label{Eq:P_Sk_Prod}
\end{equation}
Without loss of generality, we assume that $\mathbf{m} = \mathbf{0}$ (this can always be done by setting $\mathbf{x}_i - \mathbf{m} \rightarrow \mathbf{x}_i$). Then:
\begin{equation}
P(\mathcal{S}_p|c_p) = \frac{1}{ (2\pi)^{\frac{N_p D}{2}} |\mathbf{\Phi}_p|^{\frac{N_p}{2}}} exp\left( -\frac{1}{2} Tr(\mathbf{\Phi}_p^{-1} \mathbf{S}_p) \right)
\end{equation}
\begin{eqnarray}
P(\mathcal{S}_k|c_n) &=& \frac{1}{M^{-\frac{D}{2}}(2\pi)^{\frac{M D}{2}} |\mathbf{\Phi}_w|^{\frac{M-1}{2}} |\mathbf{\Phi}_n + \frac{1}{M} \mathbf{\Phi}_w|^{\frac{1}{2}}} \nonumber \\
&\cdot&exp \left( -\frac{1}{2} Tr\big((\mathbf{\Phi}_n + \frac{1}{M}\mathbf{\Phi}_w)^{-1} \mathbf{S}_n^{(k)} \big) \right) \nonumber \\
&\cdot&exp \left( -\frac{1}{2} Tr(\mathbf{\Phi}_w^{-1} \mathbf{S}_w^{(k)})\right) .
\end{eqnarray}
In the above, $\mathbf{S}_p$ is the scatter matrix of the positive class, i.e. $\mathbf{S}_p = \sum_{\mathbf{x}_i \in \mathcal{S}_p} \mathbf{x}_i \mathbf{x}_i^T$, $\mathbf{S}_w^{(k)}$ is the within-subclass scatter matrix of the $k$-th negative subclass, i.e. $\mathbf{S}_w^{(k)} = \sum_{\mathbf{x}_i \in \mathcal{S}_k} (\mathbf{x}_i - \mathbf{\bar{x}}_k)(\mathbf{x}_i - \mathbf{\bar{x}}_k)^T$ and $\mathbf{S}_n^{(k)}$ is the scatter matrix of the $k$-th subclass w.r.t. the mean vector of the positive class $\mathbf{m} = \mathbf{0}$, i.e. $\mathbf{S}_n^{(k)} = \mathbf{\bar{x}}_k \mathbf{\bar{x}}_k^T$, where $\mathbf{\bar{x}}_k$ is the mean vector of the $k$-th negative subclass, i.e. $\mathbf{\bar{x}}_k = \frac{1}{M} \sum_{\mathbf{x}_i \in \mathcal{S}_k} \mathbf{x}_i$. Derivations leading to the above results can be found in the supplementary document. Since the assignment of the negative samples to subclasses is not provided by the labels used during training, we define them by applying a clustering method (e.g. $K$-Means) on the negative class vectors.

The parameters of the proposed PCSDA are the covariance matrices $\mathbf{\Phi}_p$, $\mathbf{\Phi}_n$ and $\mathbf{\Phi}_w$ defining the data generation processes for the positive and negative classes. These parameters are estimated by maximizing the (log) probability of correct assignment of the training samples $\mathbf{x}_i, \:i=1,\dots,N$:
\begin{equation}
\mathcal{L} = \ln P(\mathcal{S}_p|c_p) + \ln P(\mathcal{S}_n|c_n),\label{Eq:LogProb}
\end{equation}
where
\begin{equation}
\ln P(\mathcal{S}_p|c_p) = C_1 - \frac{N_p}{2} \ln |\mathbf{\Phi}_p| - \frac{1}{2} Tr(\mathbf{\Phi}_p^{-1} \mathbf{S}_p)\label{Eq:PSp}
\end{equation}
and
\begin{eqnarray}
\ln P(\mathcal{S}_n|c_n) &=& \sum_{k=1}^K \ln P(\mathcal{S}_k|c_n) = C_2 - \frac{N_n-K}{2} \ln |\mathbf{\Phi}_w| \nonumber \\
&-& \frac{K}{2} \ln |\mathbf{\Phi}_n + \frac{1}{M} \mathbf{\Phi}_w| - \frac{1}{2} Tr(\mathbf{\Phi}_w^{-1} \mathbf{S}_w) \nonumber \\
&-& \frac{1}{2} Tr\big((\mathbf{\Phi}_n + \frac{1}{M}\mathbf{\Phi}_w)^{-1} \mathbf{S}_n \big), \label{Eq:PSn}
\end{eqnarray}
where $\mathbf{S}_n$ and $\mathbf{S}_w$ are the total scatter matrices of the negative subclasses, i.e., $\mathbf{S}_w = \sum_{k=1}^K \mathbf{S}_w^{(k)}$ and $\mathbf{S}_n = \sum_{k=1}^{K} \mathbf{S}_n^{(k)}$.

By substituting (\ref{Eq:PSp}) and (\ref{Eq:PSn}) in (\ref{Eq:LogProb}) the optimization problem takes the form:
\begin{eqnarray}
\mathcal{L} &=& C_3 - \frac{N_p}{2} \ln |\mathbf{\Phi}_p| - \frac{1}{2} Tr(\mathbf{\Phi}_p^{-1} \mathbf{S}_p) \nonumber \\
&-& \frac{1}{2} Tr(\mathbf{\Phi}_w^{-1} \mathbf{S}_w) - \frac{1}{2} Tr\big((\mathbf{\Phi}_n + \frac{1}{M}\mathbf{\Phi}_w)^{-1} \mathbf{S}_n \big) \nonumber \\
&-& \frac{N_n-K}{2} \ln |\mathbf{\Phi}_w| - \frac{K}{2} \ln |\mathbf{\Phi}_n + \frac{1}{M} \mathbf{\Phi}_w|.
\end{eqnarray}
The saddle points of $\mathcal{L}$ with respect to $\mathbf{\Phi}_p$, $\mathbf{\Phi}_n$, $\mathbf{\Phi}_w$ lead to:
\begin{eqnarray}
\frac{\vartheta \mathcal{L}}{\vartheta \mathbf{\Phi}_p} = \mathbf{0} &\Rightarrow& \mathbf{\Phi}_p = \frac{1}{N_p} \mathbf{S}_p \label{Eq:Phi_p}\\
\frac{\vartheta \mathcal{L}}{\vartheta \mathbf{\Phi}_n} = \mathbf{0} &\Rightarrow& \mathbf{\Phi}_n + \frac{1}{M} \mathbf{\Phi}_w =  \frac{1}{K}\mathbf{S}_n \label{Eq:Phi_n} \\
\frac{\vartheta \mathcal{L}}{\vartheta \mathbf{\Phi}_w} = \mathbf{0} &\Rightarrow& \mathbf{\Phi}_w = \frac{1}{N_n-K} \mathbf{S}_w. \label{Eq:Phi_w}
\end{eqnarray}
Note that when $N_n=K$ (\ref{Eq:Phi_w}) takes an indeterminant form (i.e. $0/0$). However, in this case each negative subclass is formed by one sample and by definition $\mathbf{\Phi}_w = \mathbf{0}$. This is discussed more in subsection \ref{SS:Extensions}. Combining (\ref{Eq:Phi_n}) and (\ref{Eq:Phi_w}) we get:
\begin{equation}
\mathbf{\Phi}_n = \frac{1}{K}\mathbf{S}_n - \frac{1}{M (N_n-K)} \mathbf{S}_w.\label{Eq:Phi_n2}
\end{equation}
By combining (\ref{Eq:Phi_w}) and (\ref{Eq:Phi_n2}) we can define the following matrix:
\begin{equation}
\mathbf{\Phi}_O = \mathbf{\Phi}_n + \mathbf{\Phi}_w = \frac{1}{K} \mathbf{S}_n + \frac{1}{N_n} \mathbf{S}_w,  \label{Eq:Phi_o}
\end{equation}
Thus, as can be seen from (\ref{Eq:Phi_p}), (\ref{Eq:Phi_w}), (\ref{Eq:Phi_n2}) and (\ref{Eq:Phi_o}), the parameters of PCSDA can be calculated using $\mathbf{S}_p$, $\mathbf{S}_n$ and $\mathbf{S}_w$ defined on the training vectors $\mathbf{x}_i, \:i=1,\dots,N$.

Using $\mathbf{S}_p$, $\mathbf{S}_n$ and $\mathbf{S}_w$ a data transformation $\mathbf{W} \in \mathbb{R}^{D \times D}$ can be obtained by optimizing for:
\begin{equation}
\mathbf{S}_n \mathbf{w} = \lambda \left( \mathbf{S}_p + \mathbf{S}_w \right) \mathbf{w}. \label{Eq:PnPi_eigproblem}
\end{equation}
Problem (\ref{Eq:PnPi_eigproblem}) corresponds to the generalized eigen-analysis problem of the matrices $\mathbf{S}_n$ and $\mathbf{S}_I = \mathbf{S}_p + \mathbf{S}_w$. Since the rank of $\mathbf{S}_n$ is $K$, the dimensionality of the obtained subspace is restricted to $d \le \textrm{min}(D, K)$. Eigenvectors in PCSDA define a data transformation that minimizes the intra-class variance of the positive class and the intra-cluster variance of the negative subclasses, while at the same time maps the means of the negative subclasses as far as possible from the positive class mean. Since the above-described property is optimized by treating (\ref{Eq:PnPi_eigproblem}) as maximization problems, the eigenvectors forming $\mathbf{W}$ are sorted in a decreasing order of the corresponding eigenvalues.

After obtaining $\mathbf{W}$, the intra-class and out-of-class covariance matrices of the transformed data $\mathbf{z}_i = \mathbf{W}^T \mathbf{x}_i$ are given by:
\begin{equation}
\tilde{\mathbf{\Phi}}_p = \mathbf{W}^T \mathbf{\Phi}_p \mathbf{W} \:\:\:\:\:\: \textrm{and} \:\:\:\:\:\: \tilde{\mathbf{\Phi}}_O = \mathbf{W}^T \mathbf{\Phi}_O \mathbf{W}. \label{Eq:tilde_Phi_matrices}
\end{equation}
The process followed in the training phase of PCSDA is illustrated in Algorithm \ref{Alg:TrainingPhase}.

In the above we set the assumption that the number of samples forming the negative subclasses is equal. This assumption can be relaxed following one of the following approaches. After assigning all negative samples to subclasses and calculating the negative subclass distributions, one can sample $M$ vectors from each distribution. Alternatively, one can calculate the total within-subclass matrix of the negative class by $\mathbf{S}_w = \sum_{k=1}^K \mathbf{S}_w^{(k)}$. The latter approach is commonly used in multi-class discriminant analysis variants \cite{iosifidis2013optimal}. Note that for the model's parameters calculation, only the overall scatter matrices $\mathbf{S}_p$, $\mathbf{S}_w$ and $\mathbf{S}_n$ are used.

\subsection{Test phase}\label{SS:TestPhase}
After the estimation of the model's parameters, a new sample $\mathbf{x}^{*}$ can be evaluated. The posterior probabilities of the positive and negative classes are given by:
\begin{equation}
P(c_p|\mathbf{x}^{*}) = \frac{ p(\mathbf{x}^{*}|c_p) P(c_p) }{p(\mathbf{x}^{*})}\label{Eq:Posterior_Cp}
\end{equation}
\begin{equation}
P(c_n|\mathbf{x}^{*}) = \frac{ p(\mathbf{x}^{*}|c_n) P(c_n) }{p(\mathbf{x}^{*})}. \label{Eq:Posterior_Cn}
\end{equation}
The a priori class probabilities $P(c_p)$ and $P(c_n)$ can be calculated by the proportion of the positive and negative samples in the training phase, i.e. $P(c_p) = N_p / N$ and $P(c_n) = N_n / N$. Depending on the problem at hand, it may be sometimes preferable to consider equiprobable classes, i.e. $P(c_p) = P(c_n) = 1/2$, leading to the maximum likelihood classification case. The class-conditional probabilities of the transformed sample $\mathbf{z}^{*} = \mathbf{W}^T \mathbf{x}^{*}$ are given by:
\begin{eqnarray}
p(\mathbf{x}^{*}|c_p) &=& \frac{1}{ (2\pi)^{\frac{D}{2}} |\mathbf{\Phi}_p|^{\frac{1}{2}}} exp\left( -\frac{1}{2} \mathbf{x}^{*T} \mathbf{\Phi}_p^{-1} \mathbf{x}^{*} \right) \nonumber \\
&=& \frac{1}{ (2\pi)^{\frac{D}{2}} |\tilde{\mathbf{\Phi}}_p|^{\frac{1}{2}}} exp\left( -\frac{1}{2} \mathbf{z}^{*T}\tilde{\mathbf{\Phi}}_p^{-1}\mathbf{z}^{*} \right)\label{Eq:Conditional_Cp}
\end{eqnarray}
and
\begin{eqnarray}
p(\mathbf{x}^{*}|c_n) &=& \frac{1}{(2\pi)^{\frac{D}{2}} |\mathbf{\Phi}_O|^{\frac{1}{2}} } exp\left( -\frac{1}{2} \mathbf{x}^{*T} \mathbf{\Phi}_O^{-1} \mathbf{x}^{*} \right) \nonumber \\
&=& \frac{1}{(2\pi)^{\frac{D}{2}} |\tilde{\mathbf{\Phi}}_O|^{\frac{1}{2}} } exp\left( -\frac{1}{2} \mathbf{z}^{*T} \tilde{\mathbf{\Phi}}_O^{-1} \mathbf{z}^{*} \right). \label{Eq:Conditional_Cn}
\end{eqnarray}
In the above, we used the orthogonal property of the matrix $\mathbf{W}$. In the case where we want $\mathbf{z}^{*} \in \mathbb{R}^d, \:d < D$, we keep the dimensions of $\mathbf{z}^{*}$ corresponding to the first $d$ columns of $\mathbf{W}$.

Combining (\ref{Eq:Posterior_Cp})-(\ref{Eq:Conditional_Cn}) the ratio of class posterior probabilities is equal to:
\begin{equation}
\lambda(\mathbf{z}^{*}) = \frac{P(c_p) \:|\tilde{\mathbf{\Phi}}_O|^{\frac{1}{2}} }{ P(c_n) |\tilde{\mathbf{\Phi}}_p|^{\frac{1}{2}}} \:\:\frac{exp\left( -\frac{1}{2} \mathbf{z}^{*T}\tilde{\mathbf{\Phi}}_p^{-1}\mathbf{z}^{*} \right)}{exp\left( -\frac{1}{2} \mathbf{z}^{*T} \tilde{\mathbf{\Phi}}_O^{-1} \mathbf{z}^{*} \right)}. \label{Eq:Lambda}
\end{equation}
$\lambda(\mathbf{z}^{*})$ can be used to classify $\mathbf{z}^{*}$ to the positive class when $\lambda(\mathbf{z}^{*}) > 1$ and to the negative class otherwise. Alternatively, (\ref{Eq:Lambda}) can be also used to define the classification rule:
\begin{eqnarray}
g(\mathbf{z}^{*}) &=& \ln P(c_p) - \ln P(c_n) \nonumber \\
&+& \frac{1}{2} \ln |\tilde{\mathbf{\Phi}}_O| - \frac{1}{2} \ln |\tilde{\mathbf{\Phi}}_p| \nonumber \\
 &-&\frac{1}{2} \mathbf{z}^{*T}\tilde{\mathbf{\Phi}}_p^{-1}\mathbf{z}^{*} + \frac{1}{2} \mathbf{z}^{*T}\tilde{\mathbf{\Phi}}_O^{-1}\mathbf{z}^{*} \label{Eq:classifier}
\end{eqnarray}
classifying $\mathbf{z}^{*}$ to the positive class if $g(\mathbf{z}^{*}) > 0$ and to the negative class otherwise.

In class-specific ranking settings one can follow the process applied when using the standard CSDA approach. First, test vectors $\mathbf{x}^{*}_j$ are mapped to the discriminant subspace $\mathbf{z}^{*}_j = \mathbf{W}^T \mathbf{x}^{*}_j$ and $\mathbf{z}^{*}_j, \:i=1,\dots,N$ are obtained. Then, $\mathbf{z}^{*}_j$'s are ordered based on their distance w.r.t. the positive class mean $d_j = \|\mathbf{z}^{*}_j - \mbox{\boldmath$\mu$}\|_2$, where $\mbox{\boldmath$\mu$}$ is the mean of positive training samples in $\mathbb{R}^d$. The process followed in the test phase of PCSDA is illustrated in Algorithm \ref{Alg:TestPhase}.
\begin{algorithm}[!t]
\SetAlgoLined
\caption{PCSDA: training phase}\label{Alg:TrainingPhase}
\KwData{$K$, $d$ and training data $\{\mathbf{x}_i,l_i\}_{i=1,\dots,N}$}
\KwResult{$\tilde{\mathbf{\Phi}}_p$, $\tilde{\mathbf{\Phi}}_O$, $\mathbf{W}$, $\mathbf{m}$, $P(c_p)$, $P(c_n)$}
 \If{$K < N_n$}{
 	Cluster $\{\mathbf{x}_i\}_{l_i \neq 1}$ to form $\mathcal{S}_k, \:k=1,\dots,K$ 
 }
 Calculate $\mathbf{m}$, $\mathbf{S}_p$, $\mathbf{S}_n$ and $\mathbf{S}_w$\\
 Calculate $\mathbf{\Phi}_p$ from (\ref{Eq:Phi_p}) and $\mathbf{\Phi}_O$ from (\ref{Eq:Phi_o})\\
 Calculate $\mathbf{W} = [\mathbf{w}_1,\dots,\mathbf{w}_d]$ from (\ref{Eq:PnPi_eigproblem})\\
 Calculate $\tilde{\mathbf{\Phi}}_p$ and $\tilde{\mathbf{\Phi}}_O$ from (\ref{Eq:tilde_Phi_matrices})
\end{algorithm}
\begin{algorithm}[!t]
\SetAlgoLined
\caption{PCSDA: test phase}\label{Alg:TestPhase}
\KwData{$\tilde{\mathbf{\Phi}}_p$, $\tilde{\mathbf{\Phi}}_O$, $\mathbf{W}$, $\mathbf{m}$, $P(c_p)$, $P(c_n)$ and test data $\mathbf{x}^{*}_i, \:i=1,\dots,M$}
\KwResult{Predicted labels $l_i, \:i=1,\dots,M$ or ranking order $o_i,\:i=1,\dots,M$}
 Calculate $\mbox{\boldmath$\mu$} = \mathbf{W}^T \mathbf{m}$ and $\mathbf{z}^{*}_i = \mathbf{W}^T \mathbf{x}^{*}_i, \:i=1,\dots,M$\\
 \eIf{Classification}{
   \For{i=1:M}{
   Calculate $\tilde{\mathbf{z}}^{*}_i = \mathbf{z}^{*}_i - \mbox{\boldmath$\mu$}$\\
   \eIf{$g(\tilde{\mathbf{z}}^{*}_i) \ge 0$}{ 
   		$l_i$ = 1\;
   		}{ 
   		$l_i$ = 0\;
   		}
   }
   }{
   Calculate $d_i = \|\mbox{\boldmath$\mu$} - \mathbf{z}^{*}_i\|_2, \:i=1,\dots,M$\\
   Sort $d_i$'s to obtain the order $o_i, \:i=1,\dots,M$
  }
\end{algorithm}

\subsection{Spectral Regression solution of PCSDA}\label{SS:SpectralRegressionPCSDA}
We further show in the following that PCSDA can be efficiently solved by following a spectral regression based process. Let us denote by $\mathbf{1}_p \in \mathbb{R}^N$ a vector having ones in the elements corresponding to the positive training vectors and zeros elsewhere. In the same manner, we define the vector $\mathbf{1}_k \in \mathbb{R}^N$ as a vector having ones in the elements corresponding to the negative samples belonging to the $k$-th subclass and zeros elsewhere. We also define by $\mathbf{J}_p \in \mathbb{R}^{N \times N}$ and $\mathbf{J}_k \in \mathbb{R}^{N \times N}$ the diagonal matrices having in their diagonal the vectors $\mathbf{1}_p$ and $\mathbf{1}_k$, respectively. Then, $\mathbf{S}_n$ and $\mathbf{S}_I$ can be expressed as:
\begin{equation}
\mathbf{S}_n = \sum_{k=1}^{K} (\mathbf{\bar{x}}_k - \mathbf{m}) (\mathbf{\bar{x}}_k - \mathbf{m})^T = \mathbf{X} \mathbf{L}_n \mathbf{X}^T, \label{Eq:Ln}
\end{equation}
\begin{equation}
\mathbf{S}_I = \mathbf{S}_p + \mathbf{S}_w = \mathbf{X} \mathbf{L}_I \mathbf{X}^T \label{Eq:Lw}
\end{equation}
where
\begin{equation}\label{Eq:L_n_eq}
\mathbf{L}_n = \sum_{k=1}^{K} \left( \frac{1}{N_k^2} \mathbf{1}_k \mathbf{1}_k^T - \frac{1}{N_k N_p} (\mathbf{1}_k \mathbf{1}_p^T + \mathbf{1}_p \mathbf{1}_k^T) + \frac{1}{N_p^2} \mathbf{1}_p \mathbf{1}_p^T \right)
\end{equation}
\begin{equation}\label{Eq:L_I_eq}
\mathbf{L}_I = \left(\mathbf{J}_p - \frac{1}{N_p} \mathbf{1}_p \mathbf{1}_p^T \right)  + \sum_{k=1}^{K} \left(\mathbf{J}_k - \frac{1}{N_k} \mathbf{1}_k \mathbf{1}_k^T \right).
\end{equation}
Substituting (\ref{Eq:Ln}) and (\ref{Eq:Lw}) in (\ref{Eq:PnPi_eigproblem}) and setting $\mathbf{v} = \mathbf{X}^T \mathbf{w}$:
\begin{equation}
\mathbf{L}_n \mathbf{v} = \lambda \mathbf{L}_I \mathbf{v}. \label{Eq:EigProbLnLw}
\end{equation}
Thus, the vectors $\mathbf{v}$ are the eigenvectors of the matrix $\mathbf{L} = \tilde{\mathbf{L}}^{-1}_w \mathbf{L}_n$, where $\tilde{\mathbf{L}}_w$ is a regularized version of $\mathbf{L}_w$, i.e. $\tilde{\mathbf{L}}_w = \mathbf{L}_w + \epsilon \mathbf{I}$ where $\epsilon > 0$. It can be shown that the matrix $\mathbf{L}$ is a block matrix formed by blocks the elements of which are indicated by the class labels of the positive data and the cluster labels of the negative data. The top $K$ eigenvectors of it are also formed by blocks indicated by the same labels. Thus, they can be defined without the need of solving the problem (\ref{Eq:EigProbLnLw}), in a similar manner as in \cite{cai2007spectral,iosifidis2017cskdaRev}. After the determination of the matrix $\mathbf{V} = [\mathbf{v}_1,\dots,\mathbf{v}_K]$, the solution of (\ref{Eq:PnPi_eigproblem}) is given by $\mathbf{W} = \left(\mathbf{X}^T\right)^{\dag}\mathbf{V}$, where $\left(\mathbf{X}^T\right)^{\dag}$ is the pseudo-inverse of $\mathbf{X}^T$.

\subsection{CSDA variants under the probabilistic model}\label{SS:Extensions}
A special case of the PCSDA can be obtained by setting the assumption that each negative sample forms a negative subclass, i.e. $K = N_n$ and $M = 1$. In that case $\mathbf{\Phi}_w = \mathbf{0}$, $\mathbf{\Phi}_O = \mathbf{\Phi}_n$, the negative samples are drawn from a distribution $P(\mathbf{x}) \sim N(\cdot | \mathbf{m},\mathbf{\Phi}_n)$, and $\mathbf{W}$ is calculated by solving for $\mathbf{S}_n \mathbf{w} = \tilde{\lambda} \mathbf{S}_p \mathbf{w}$, i.e., we obtain the class discrimination definition of CSDA. The Spectral Regression-based solutions of CSDA in \cite{arashloo2014csksr,iosifidis2015CSRDA,iosifidis2016sucskda} and the low-rank regression solution of \cite{iosifidis2017cskdaRev} use the same class discrimination criterion and, thus, correspond to the same setting of PCSDA. Since the discrimination criterion used in CSDA is a special case of the proposed probabilistic model, all the above-mentioned methods can be extended to perform classification using $g(\cdot)$ in (\ref{Eq:classifier}). Jointly optimizing the data projection and the classification rule can be considered a big advantage.

\subsection{Class-specific vs. Contrastive Learning}\label{SS:ContrustiveLearning}
As can be seen from Eqs. (\ref{Eq:CSKDA_J2}), (\ref{Eq:S_n_matrix}) and (\ref{Eq:S_p_matrix}), the Class-specific Discriminant Analysis criterion has similarities with Contrastive Learning based on triplet loss \cite{weinberger2009distance}, which is defined for multi-class problems. That is, in its generic form, it optimizes for class discrimination by using three samples, i.e. the sample of interest, a sample belonging to the same class with the sample of interest (defining the same-label pair), and a sample belonging to a different class (defining the different-label pair). Thus, it can be modeled by a graph-based criterion, as was shown in \cite{hedegaard2020dage}. The class-specific learning approaches are defined on binary problems where the negative class is formed data coming from multiple classes, but are assigned the same (negative) label. All the class-specific discriminant analysis approaches (including the proposed one) are defined using variance criteria. They optimize for class discrimination by using the mean of the positive class, a positive sample, and a negative sample. Thus, they can be defined as graph-based methods having the Laplacian matrices $\mathbf{L}_n$ and $\mathbf{L}_I$ in Eqs. (\ref{Eq:L_n_eq}) and (\ref{Eq:L_I_eq}). 

The adoption of class-specific learning has two advantages compared to contrastive learning when class-specific subspace learning is considered:
\begin{itemize}
    \item Subspace learning based on triplet loss optimizes for two objectives:
    \begin{enumerate}
        \item Given a positive vector of interest, it should be mapped close to another positive vector (same-label pair), and far away from a negative vector (different-label pair).
        \item Given a negative vector of interest, it should be mapped close to another negative vector (same-label pair), and far away from a positive vector (different-label pair).
    \end{enumerate}
    While objective $1$ is desirable in class-specific problems, optimizing for objective $2$ can be destructive. Optimizing for both objectives is good for multi-class problems when assuming class unimodality, like in the case of LDA.
    
    \item Class-specific learning defined by a variance-based criterion can lead to fast solutions. This is due to that the graph Laplacian matrices $\mathbf{L}_n$ and $\mathbf{L}_I$ do not need to be calculated and stored in the memory. Instead, their eigenvectors can be efficiently obtained and highly reduce the memory and computational cost. This is described in Section \ref{SS:SpectralRegressionPCSDA}.
\end{itemize}

\subsection{Non-linear PCSDA}\label{SS:kernelPCSDA}
In the above analysis we considered the linear class-specific subspace learning case. In order to non-linearly map $\mathbf{x}_i \in \mathbb{R}^D$ to $\mathbf{z}_i \in \mathbb{R}^d$ traditional kernel-based learning methods perform a non-linear mapping of the input space $\mathbb{R}^D$ to the feature space $\mathcal{F}$ using a function $\phi(\cdot)$, i.e. $\mathbf{x}_i \in \mathbb{R}^D \rightarrow \phi(\mathbf{x}_i) \in \mathcal{F}$. Then, linear class-specific projections are defined by using the training data in $\mathcal{F}$. Since the dimensionality of $\mathcal{F}$ is arbitrary (virtually infinite), the data representations in $\mathcal{F}$ cannot be calculated. Traditional kernel-based learning methods address this issue by exploiting the Representer theorem and the non-linear mapping is implicitly performed using the kernel function encoding dot products in the feature space, i.e. $\kappa(\mathbf{x}_i,\mathbf{x}_j) = \phi(\mathbf{x}_i)^T \phi(\mathbf{x}_j)$ \cite{Schlkopf2001}.

As has been shown in \cite{kwak2013npt} the effective dimensionality of the kernel space $\mathcal{F}$ is at most equal to $L = \min(D,N)$ and, thus, an explicit non-linear mapping $\mathbf{x}_i \in \mathbb{R}^D \rightarrow \mathbf{\phi}_i \in \mathbb{R}^L$ can be calculated such that $\kappa(\mathbf{x}_i,\mathbf{x}_j) = \mathbf{\phi}_i^T \mathbf{\phi}_j$. This is achieved by using $\mathbf{\Phi} = \mathbf{\Sigma}^{\frac{1}{2}} \mathbf{U}^T$, where $\mathbf{U}$ and $\mathbf{\Sigma}$ contain the eigenvectors and eigenvavlues of the kernel matrix $\mathbf{K} \in \mathbb{R}^{N \times N}$ \cite{kwak2017IncrNPT}. Thus, extension of PCSDA to the non-linear (kernel) case can be readily obtained by applying the above-described linear PCSDA on the vectors $\mathbf{\phi}_i, \:i=1,\dots,N$. Here we need to mention that, since the proposed method exploits kernel-based data representations, it inherits drawbacks of kernel methods related to their high computational and memory costs when applied to large-scale problems. For the cases where the size of training set is prohibitive for applying kernel-based discriminant learning, the Nystr\"{o}m-based kernel subspace learning method of \cite{iosifidis2016anpt} or nonlinear data mappings based on randomized features, as proposed in \cite{rahimi2007rfr}, can be used. We should also note that the application of $K$-Means using in $\mathbb{R}^L$ corresponds to the application of kernel $K$-Means in the original space $\mathbb{R}^D$.

\subsection{Time complexity}\label{SS:Complexity}
The time complexity of PCSDA is as follows \cite{golub1996matrix}:
\begin{itemize}
    \item $K$-Means application for determining the subclasses of the negative class having time complexity of $O(\eta K N_n D)$, where $\eta$ is the number of iterations until convergence,
    \item Calculation of the scatter matrices in (\ref{Eq:PnPi_eigproblem}) and solution of the generalized eigen-analysis problem having a time complexity of $O(D^3 + D^2N)$,
\end{itemize}
Keeping the high-order terms, the overall time complexity is of $O(D^3 + D^2 N)$. 
\begin{table}[]
\begin{center}
\caption{Datasets information.}\label{tbl:DataInfo}
\resizebox{0.8\linewidth}{!}{
\begin{tabular}{|l|c|c|c|} \cline{1-4}
Dataset                                    &  $D$  & $\#$Samples  & $\#$Classes \\ \hline
Jaffe      \cite{Lyons1998coding}          & 1200  &    210       &     7       \\ \hline
Kanade     \cite{kanade2000comprehensive}  & 1200  &    245       &     7       \\ \hline
BU         \cite{Yin2006facial}            & 1200  &    700       &     7       \\ \hline
YALE       \cite{lee2005acquiring}         & 1200  &   2432       &    38       \\ \hline
AR         \cite{martinez2001pca}          &  1200 &   2600       &   100       \\ \hline
15 scenes  \cite{lazebnik2006beyond}       &  512  &   4485       &    15       \\ \hline
OptDigits  \cite{Dua2019UCI}               &   54  &   5620       &    10       \\ \hline 
Caltech101 \cite{quattonni2009recognizing} &  512  &   9145       &   102       \\ \hline  
\end{tabular}}
\end{center}
\end{table}
\begin{table*}
\begin{center}
\caption{Performance on retrieval problems.}\label{tbl:Ranking}
\resizebox{\linewidth}{!}{
\begin{tabular}{|l|c|c||c|c|c|c|} \cline{2-7}
\multicolumn{1}{c|}{ }          &          RR         &          SVR        &        PCA        &         LDA         &          PCSDA-1      &  PCSDA-K   \\ \hline  
      Jaffe  &  0.9024 (0.0935)  &  0.8954 (0.1179)  &  0.3010 (0.1383)   &  0.9052 (0.0990)  &  0.9099 (0.0953)  &  0.7591 (0.1728)    \\ \hline  
     Kanade  &  0.6523 (0.2635)  &  0.6525 (0.2625)  &  0.2930 (0.1139)   &  0.6336 (0.2476)  &  0.6621 (0.2573)  &  0.4196 (0.2629)    \\ \hline  
		 BU  &  0.6848 (0.2279)  &  0.6976 (0.2346)  &  0.2903 (0.1455)   &  0.6491 (0.2099)  &  0.6840 (0.1940)  &  0.6520 (0.1902)    \\ \hline  
       YALE  &  0.9874 (0.0250)  &  0.9883 (0.0231)  &  0.5580 (0.0410)   &  0.9864 (0.0251)  &  0.9864 (0.0251)  &  0.9911 (0.0189)    \\ \hline  
         AR  &  0.9990 (0.0057)  &  0.9976 (0.0109)  &  0.8642 (0.0113)   &  0.9988 (0.0061)  &  0.9986 (0.0071)  &  0.9959 (0.0194)    \\ \hline  
   15 scenes &  0.9340 (0.0409)  &  0.9403 (0.0387)  &  0.6161 (0.1805)   &  0.9160 (0.0587)  &  0.9162 (0.0602)  &  0.9127 (0.0549)    \\ \hline  
  OptDigits  &  0.9968 (0.0036)  &  0.9969 (0.0044)  &  0.8337 (0.1352)   &  0.9974 (0.0033)  &  0.9970 (0.0034)  &  0.9945 (0.0064)    \\ \hline  
 Caltech101  &  0.8708 (0.1383)  &  0.8792 (0.1301)  &  0.2239 (0.2000)   &  0.8485 (0.1513)  &  0.8467 (0.1514)  &  0.8523 (0.1486)    \\ \hline
\end{tabular}}
\end{center}
\end{table*}
\begin{table*}
\begin{center}
\caption{Performance on classification problems.}\label{tbl:Classification}
\resizebox{\linewidth}{!}{
\begin{tabular}{|l|c|c||c|c|c|c|} \cline{2-7}
\multicolumn{1}{c|}{ }          &          RR         &          SVM        &      PCA+SVM      &         LDA+NCC         &          PCSDA-1      &  PCSDA-K   \\ \hline  
      Jaffe  &  0.6750 (0.1032)  &  0.6714 (0.1225)  &  0.2809 (0.0542)   &  0.7049 (0.1299)  &  0.6261 (0.1262)  &  0.5242 (0.1249)    \\ \hline  
     Kanade  &  0.4001 (0.1520)  &  0.4122 (0.1271)  &  0.2597 (0.0464)   &  0.5194 (0.2677)  &  0.4546 (0.2427)  &  0.3861 (0.1894)    \\ \hline  
		 BU  &  0.4463 (0.0819)  &  0.4562 (0.1391)  &  0.2592 (0.0180)   &  0.5632 (0.1706)  &  0.5731 (0.1288)  &  0.5431 (0.1793)    \\ \hline  
       YALE  &  0.9369 (0.0090)  &  0.9396 (0.0851)  &  0.2715 (0.0851)   &  0.9517 (0.0208)  &  0.9409 (0.0214)  &  0.9611 (0.0222)    \\ \hline  
         AR  &  0.9664 (0.0352)  &  0.9683 (0.0343)  &  0.7346 (0.0394)   &  0.9760 (0.0452)  &  0.9674 (0.0481)  &  0.9516 (0.0629)    \\ \hline  
   15 scenes &  0.8997 (0.0854)  &  0.9005 (0.0517)  &  0.5365 (0.1507)   &  0.8993 (0.0539)  &  0.8573 (0.0654)  &  0.8130 (0.1275)    \\ \hline  
  OptDigits  &  0.9875 (0.0900)  &  0.9881 (0.0082)  &  0.8985 (0.0706)   &  0.9899 (0.0077)  &  0.9569 (0.0141)  &  0.9826 (0.0074)    \\ \hline  
 Caltech101  &  0.8130 (0.0971)  &  0.8140 (0.0807)  &  0.6042 (0.0394)   &  0.8075 (0.1448)  &  0.7605 (0.1829)  &  0.7663 (0.1955)    \\ \hline
\end{tabular}}
\end{center}
\end{table*}

Considering the kernel-based version of PCSDA, its time complexity is as follows:
\begin{itemize}
    \item Kernel matrix $\mathbf{K}$ calculation having a time complexity of $O(DN^2)$.
    \item Application of the method \cite{kwak2013npt}, involving the eigen-decomposition of $\mathbf{K}$, having a time complexity of $O(N^3)$.
    \item $K$-Means application for determining the negative class subclasses having complexity of the order of $O(\eta K N_n N)$.
    \item Calculation of the scatter matrices in (\ref{Eq:PnPi_eigproblem}) and solution of the generalized eigen-analysis problem having a time complexity of $O(N^3 + 2 N^2)$
\end{itemize}
Thus, the time complexity of the kernel version of PCSDA is $O(N^3)$, which is the case of a generic kernel-based subspace learning method. As noted in subsection \ref{SS:kernelPCSDA}, for applying nonlinear PCSDA on big datasets, the approximate kernel subspace learning method in \cite{iosifidis2016anpt} or nonlinear data mappings based on randomized features, as proposed in \cite{rahimi2007rfr}, can be used leading to $L \ll N$ and highly reducing the overall time complexity to $O(L^3 + LN^2)$.

Finally, considering the Spectral Regression-based solution described in Section \ref{SS:SpectralRegressionPCSDA}, the time complexity of a generic solution is the same as those described above, but efficient solutions are possible by using efficient decomposition methods, like Cholesky decomposition as in \cite{iosifidis2017cskdaRev}.

\section{Experiments}\label{S:Experiments}
In this Section we provide experimental results illustrating the performance of the proposed PCSDA method. We used eight datasets in our experiments, details of which are illustrated in Table \ref{tbl:DataInfo}. For BU, Jaffe, Kanade, YALE and AR facial image datasets we used the vectorized pixel intensity values for representing the images. For the $15$ scenes and Caltech101 datasets we used deep features generated by average pooling over spatial dimension of the last convolution layer of VGG network \cite{simonyan2014very} trained on ILSVRC2012 database. For optDigits dataset we used the raw features provided by the UCI repository \cite{Dua2019UCI}.

On each experiment, we formed class-specific ranking and classification problems for each class using the class data as positive samples and the data of the remaining classes as negative samples. In all the experiments the data is non-linearly mapped to the subspace of the kernel space (as discussed in Section \ref{SS:kernelPCSDA}). We used the RBF kernel function and set the value of $\sigma$ equal to the mean pair-wise distance value between the positive training vectors. For the small datasets, we used the method in \cite{kwak2017IncrNPT} by keeping the eigenvectors corresponding to all positive eigenvalues, while for the large datasets we used the method in \cite{iosifidis2016anpt} by setting the dimensionality of the resulted kernel subspace to $L = 1000$. On each class-specific problem, we ran five experiments by randomly selecting $70\%$ of the positive and negative classes for training and the rest $30\%$ for testing and measure the average performance value. The hyper-parameter values of all methods have been optimized by applying five-fold cross-validation on the training data. Specifically, for the methods performing subspace learning, the dimensionality of the obtained subspace was selected within the range of $[1,25]$ and for Support Vector Machine (SVM), Support Vector Regression (SVR) and Ridge Regression (RR) the regularization parameter value was selected in the range of $10^r, \:r=-3,\dots,3$. We used the publicly available software implementations from \cite{pan2008liblinear} for the support vector methods. We tested two variants of the proposed PCSDA, one in which the number of negative subclasses is fixed to $K = 1$ (noted as PCSDA-1) and one in which the number of negative subclasses is automatically selected during the training process following the five-fold cross validation process from the set $K = \{5, 10, 15, 20\}$ (noted as PCSDA-K). We opt to automatically determine the number of negative subclasses (and not set $K$ equal to the number of the classes of the multi-class classification problem forming the negative class) to appropriately address possible issues of class multi-modality. Indeed, by following such an approach, negative subclasses can be formed by samples belonging to more than one classes of the original multi-class classification problem, and multiple negative subclasses can be formed by samples belonging to the same class of the original multi-class classification problem. We should note here that the case $K = 1$ is a subcase of the generic PCSDA-K.

For evaluating the performance of the methods in a ranking setting, we use the mean Average Precision (mAP) metric. For evaluating the performance of the methods in a classification setting, we use the $f_1$ score defined as $f_1 = 2 \:\frac{precision*recall}{precision + recall}$. Performance obtained for each method in a ranking and a classification setting is shown in Tables \ref{tbl:Ranking} and \ref{tbl:Classification}, respectively. The values of the Tables correspond to the mean performance value and the corresponding standard deviation over all class-specific problems of each dataset.

As can be seen in Table \ref{tbl:Ranking} for the ranking problems, the use of an unsupervised subspace learning method (PCA) leads to low performance. The use of regression models (RR and SVR) achieves high performance in most of the datasets, except Kanade and BU. The case is similar for the supervised subspace learning methods (LDA and PCSDA). It is interesting to see that in these two datasets PCSDA outperforms LDA by a margin of around $3\%$, while for the rest datasets their performance is similar. Focusing on PCSDA-1 and PCSDA-K, we can see that in most cases the use of one negative subclass seems to provide (slightly) better results, except for the Yale and Caltech101 datasets. We speculate that this is due to the homogeneity of the remaining datasets, however, we can see that the use of multiple subclasses can enhance performance for heterogeneous data. 

Table \ref{tbl:Classification} illustrates the performance of the methods in classification problems. Here we need to note that for performing classification using a subspace-based method (PCA and LDA) one needs to perform a two-step process, i.e. data projection to the subspace followed determined by the method followed by the application of a classification method. We combine PCA with SVM and LDA with Nearest Class Centroid (NCC) classifiers. The choice of LDA+NCC is based on the fact that the LDA optimization problem exploits as a class separability criterion that of NCC. Despite the fact that PCSDA is a subspace learning method, its probabilistic formulation leads to a classification rule (as detailed in Section \ref{SS:TestPhase}), and thus there is no need to perform a classification method for PCSDA. We can again see that the use of unsupervised subspace learning combined with SVM leads to low performance. Subspace learning-based classification leads to good performance, outperforming regression-based classification and SVM in most cases. Interestingly, the use of multiple negative subclasses leads to an improvement on Caltech101, Yale and OptDigits datasets of sizes $0.5\%$, $2\%$ and $2.5\%$, respectively.

\section{Conclusions}\label{S:Conclusions}
In this paper we proposed a probabilistic model for class-specific discriminant subspace learning that is able to incorporate subclass information of the negative class in the optimization criterion. The adoption of a probability-based optimization criterion for class-specific discriminant subspace learning leads to a classifier defined on the data representations in the discriminant subspace. We showed that the parameters of the probabilistic model can be obtained by applying an efficient Spectral Regression-based process. Moreover, we showed that the proposed probabilistic model includes as special cases existing class-specific discriminant analysis methods. Experimental results illustrated the performance of the proposed model, in comparison with that of related methods, in both verification and classification problems. Interesting extensions of the proposed approach would be to devise class-specific criteria for Subspace Clustering \cite{pend2016constructing,pend2018structured} and multi-view based subspace learning \cite{peng2019comic,cao2018generalized}.

\section*{Acknowledgement}
This work has received funding from the European Union’s Horizon 2020 research and innovation programme under grant agreement No 871449 (OpenDR). This publication reflects the authors’ views only. The European Commission is not responsible for any use that may be made of the information it contains.

\bibliographystyle{IEEEtran}
\bibliography{bibliography}

\includepdf[page=-]{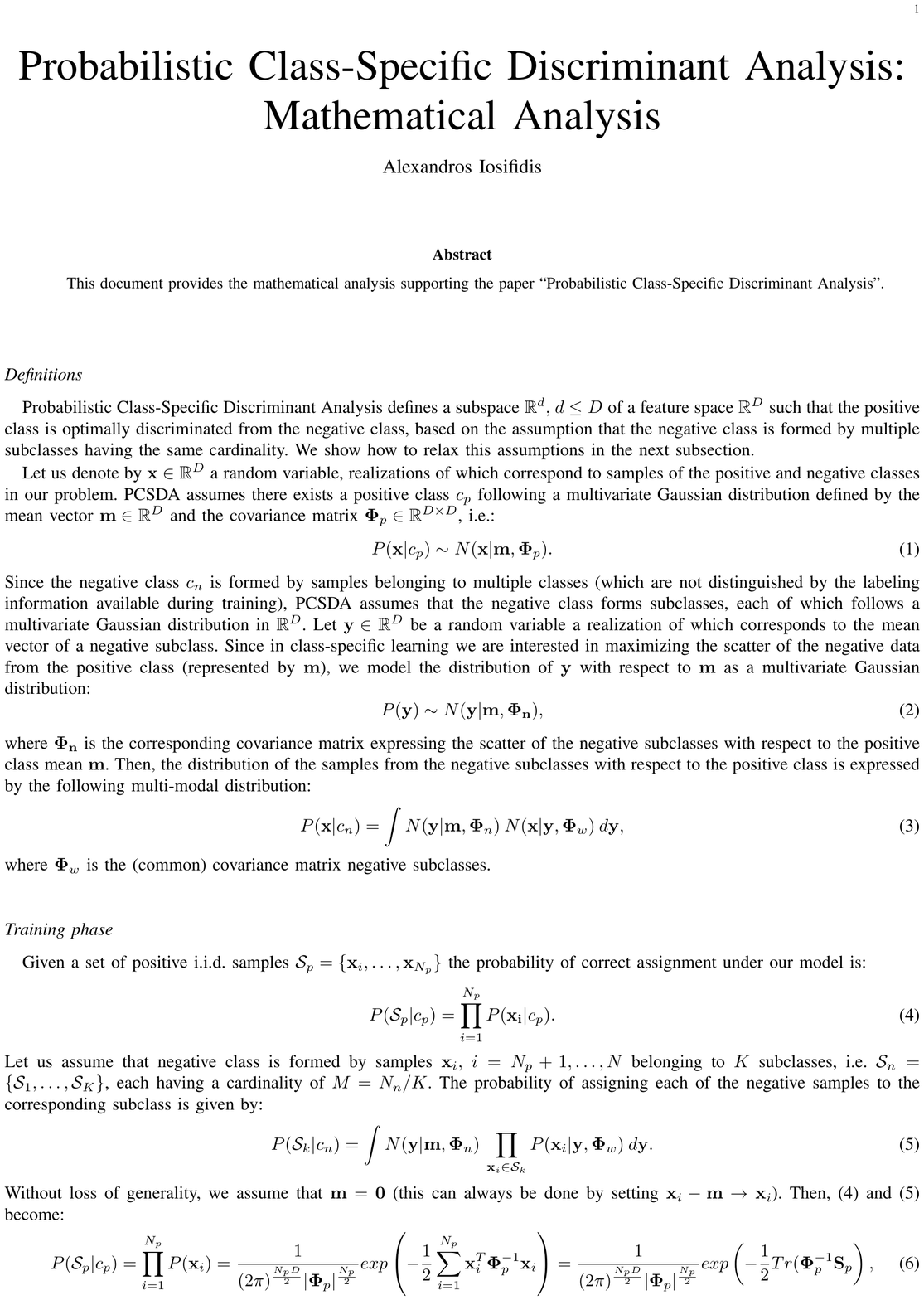}

\end{document}